# Non-Convex Rank Minimization via an Empirical Bayesian Approach

David Wipf
Visual Computing Group, Microsoft Research Asia
davidwipf@gmail.com

## Abstract

In many applications that require matrix solutions of minimal rank, the underlying cost function is non-convex leading to an intractable, NP-hard optimization problem. Consequently, the convex nuclear norm is frequently used as a surrogate penalty term for matrix rank. The problem is that in many practical scenarios there is no longer any guarantee that we can correctly estimate generative low-rank matrices of interest, theoretical special cases notwithstanding. Consequently, this paper proposes an alternative empirical Bayesian procedure build upon a variational approximation that, unlike the nuclear norm, retains the same globally minimizing point estimate as the rank function under many useful constraints. However, locally minimizing solutions are largely smoothed away via marginalization, allowing the algorithm to succeed when standard convex relaxations completely fail. While the proposed methodology is generally applicable to a wide range of low-rank applications, we focus our attention on the robust principal component analysis problem (RPCA), which involves estimating an unknown low-rank matrix with unknown sparse corruptions. Theoretical and empirical evidence are presented to show that our method is potentially superior to related MAP-based approaches, for which the convex principle component pursuit (PCP) algorithm (Candès et al., 2011) can be viewed as a special case.

## 1 INTRODUCTION

Recently there has been a surge of interest in finding low-rank decompositions of matrix-valued data subject to some problem-specific constraints (Babacan et al., 2011; Candès & Recht, 2008; Candès et al., 2011; Chandrasekaran et al., 2011; Ding et al., 2011). While the methodology proposed herein is applicable to a wide range of low-rank applications, we will focus our attention on the robust principal component analysis problem (RPCA) described in Candès et al. (2011). We begin with the observation model

$$Y = X + S + E, \qquad (1)$$

where $Y \in \mathbb{R}^{m \times n}$ is an observed data matrix, $X$ is an unknown low-rank component, $S$ is a sparse corruption matrix, and $E$ is diffuse noise, with iid elements distributed as $N(0, \lambda)$. Without loss of generality, we will assume throughout that $n \geq m$. To estimate $X$ and $S$ given $Y$, one possibility is to solve

$$\min_{X,S} \frac{1}{\lambda} \|Y - X - S\|_{\mathcal{F}}^2 + n\text{Rank}\,[X] + \|S\|_0, \qquad (2)$$

where $\|S\|_0$ denotes the matrix $\ell_0$ norm of $S$, or a count of the nonzero elements in $S$. The reason for the factor of $n$ is to ensure that the rank and sparsity terms are properly balanced, meaning that both terms range from 0 to $nm$, which reflects our balanced uncertainty regarding their relative contributions to $Y$. Unfortunately, solving (2) is problematic because the objective function is both discontinuous and non-convex. In general, the only way to guarantee that the global minimum is found is to conduct an exhaustive combinatorial search, which is intractable in all but the simplest cases.

The most common alternative, sometimes called principle component pursuit (PCP) (Candès et al., 2011), is to replace (2) with a convex surrogate such as

$$\min_{X,S} \frac{1}{\lambda} \|Y - X - S\|_{\mathcal{F}}^2 + \sqrt{n}\|X\|_* + \|S\|_1, \qquad (3)$$

where $\|X\|_*$ denotes the nuclear norm of $X$ (or the sum of its singular values). Note that the scale factor from (2) has been changed from $n$ to $\sqrt{n}$; this is an artifact of the relaxation mechanism balancing the nuclear and $\ell_1$ norms.[1] A variety of recent theoretical results stipulate when solutions of (3), particularly in the limiting case where $\lambda \to 0$ (reflecting the assumption that $E = 0$), will produce reliable estimates of $X$ and $S$ (Candès et al., 2011; Chandrasekaran et al., 2011). However, in practice these results have marginal value since they are based upon strong, typically unverifiable assumptions on the support of $S$ and the structure of $X$. In general, the allowable support of $S$ may be prohibitively small and unstructured (possibly random); related assumptions are required for the rank and structure of $X$. Thus there potentially remains a sizable gap between what can be achieved by minimizing the 'ideal' cost function (2) and the convex relaxation (3).

In Section 2, as a motivational tool we discuss a simple non-convex scheme based on a variational majorization-minimization approach for locally minimizing (2). Then in Section 3 we reinterpret this method as *maximum a posteriori* (MAP) estimation and use this perspective to design an alternative empirical Bayesian algorithm that avoids the major shortcomings of MAP estimation. Section 4 investigates analytical properties of this empirical Bayesian alternative with respect to globally and locally minimizing solutions. Later in Section 5 we compare a state-of-the-art PCP algorithm with the proposed approach on simulated data as well as a photometric stereo problem.

## 2 NON-CONVEX MAJORIZATION-MINIMIZATION

One possible alternative to (3) is to replace (2) with a non-convex yet smooth approximation that can at least be locally minimized. In the sparse estimation literature, one common substitution for the $\ell_0$ norm is the Gaussian entropy measure $\sum_i \log |s_i|$, which may also sometimes include a small regularizer to avoid taking the log of zero.[2] This can be justified (in part) by the fact that

$$\sum_i \log |s_i| \equiv \lim_{p \to 0} \frac{1}{p} \sum_i \left( |s_i|^p - 1 \right) \propto \|\boldsymbol{s}\|_0. \quad (4)$$

An analogous approximation suggests replacing the rank penalty with $\log |XX^T|$ as has been suggested for related rank minimization problems (Mohan & Fazel, 2010), since $\log |XX^T| = \sum_i \log \sigma_i$, where $\sigma_i$ are the singular values of $XX^T$. This leads to the alternative cost function

$$\min_{X,S} \frac{1}{\lambda} \|Y - X - S\|_{\mathcal{F}}^2 + n \log |XX^T| + 2 \sum_{i,j} \log |s_{ij}|. \quad (5)$$

Optimization of (5) can be accomplished by a straightforward majorization-minimization approach based upon variational bounds on the non-convex penalty terms (Jordan et al., 1999). For example, because $\log |s|$ is a concave function of $s^2$, it can be expressed using duality theory (Boyd & Vandenberghe, 2004) as the minimum of a particular set of upper-bounding lines:

$$\log s^2 = \min_{\gamma \geq 0} \frac{s^2}{\gamma} + \log \gamma - 1. \quad (6)$$

Here $\gamma$ is a non-negative variational parameter controlling the slope. Therefore, for any fixed $\gamma$ we have a strict, convex upper-bound on the concave log function. Likewise, for the rank term we can use the analogous representation (Mohan & Fazel, 2010)

$$n \log |XX^T| = \min_{\Psi \succeq 0} \mathrm{Trace} \left[ XX^T \Psi^{-1} \right] + n \log |\Psi| + C, \quad (7)$$

where $C$ is an irrelevant constant and $\Psi$ is a positive semi-definite matrix of variational parameters.[3] Combining these bounds we obtain an equivalent optimization problem

$$\min_{X,S,\Gamma \succeq 0, \Psi \succeq 0} \frac{1}{\lambda} \|Y - X - S\|_{\mathcal{F}}^2 + \sum_{ij} \left( \frac{s_{ij}^2}{\gamma_{ij}} + \log \gamma_{ij} \right)$$
$$+ \mathrm{Trace} \left[ XX^T \Psi^{-1} \right] + n \log |\Psi|, \quad (8)$$

where $\Gamma$ is a matrix of non-negative elements composed of the variational parameters corresponding to each $s_{ij}$. With $\Gamma$ and $\Psi$ fixed, (8) is quadratic in $X$ and $S$ and can be minimized in closed form via

$$\begin{aligned} \boldsymbol{x}_j &\to \Psi \left( \Psi + \bar{\Gamma}_j \right)^{-1} \boldsymbol{y}_j, \\ \boldsymbol{s}_j &\to \bar{\Gamma}_j \left( \Psi + \bar{\Gamma}_j \right)^{-1} \boldsymbol{y}_j, \quad \forall j \end{aligned} \quad (9)$$

where $\boldsymbol{y}_j$, $\boldsymbol{x}_j$, and $\boldsymbol{s}_j$ represent the $j$-th columns of $Y$, $X$, and $S$ respectively and $\bar{\Gamma}_j$ is a diagonal matrix formed from the $j$-th column of $\Gamma$. Likewise, with $X$ and $S$ fixed, $\Gamma$ and $\Psi$ can also be obtained in closed form using the updates

$$\begin{aligned} \Psi &\to \frac{1}{n} XX^T \\ \gamma_{ij} &\to s_{ij}^2, \quad \forall i, j. \end{aligned} \quad (10)$$

While local minimization of (5) is clear cut, finding global solutions can still be highly problematic just as before. Whenever any coefficient of $S$ goes to zero, or whenever the rank of $X$ is reduced, we are necessarily at a local minimum with respect to this quantity such that we can never increase the rank or a zero-valued coefficient magnitude in search of the global optimum. (This point will be examined in further detail in Section 4.) Thus the algorithm may quickly converge to one of a combinatorial number of local solutions.

---
[1] Actually, different scaling factors can be adopted to reflect different assumptions about the relative contributions of low-rank and sparse terms. But we assume throughout that no such knowledge is available.

[2] Algorithms and analysis follow through much the same regardless.

[3] If $X$ is full rank, then $\Psi$ must be positive definite.

## 3 VARIATIONAL EMPIRICAL BAYESIAN ALGORITHM

From a Bayesian perspective we can formulate (5) as a MAP estimation problem based on the distributions

$$\begin{aligned}
p(Y|X,S) &\propto \exp\left[-\frac{1}{2\lambda}\|Y - X - S\|_\mathcal{F}^2\right] \\
p(X) &\propto \frac{1}{|XX^T|^{n/2}} \\
p(S) &\propto \prod_{i,j}\frac{1}{|s_{ij}|}.
\end{aligned} \quad (11)$$

It is then transparent that solving

$$\max_{X,S} p(X,S|Y) \equiv \max_{X,S} p(Y|X,S)p(X)p(S) \quad (12)$$

is equivalent to solving (5) after an inconsequential $-2\log(\cdot)$ transformation. But as implied above, this strategy is problematic because the effective posterior is characterized by numerous spurious peaks rendering MAP estimation intractable. A more desirable approach would ignore most of these peaks and focus only on regions with significant posterior mass, regions that hopefully also include the posterior mode. One way to accomplish this involves using the bounds from (6) and (7) to construct a simple approximate posterior that reflects the mass of the original $p(X,S|Y)$ sans spurious peaks. We approach this task as follows.

From (6) and (7)) we can infer that

$$\begin{aligned}
p(S) &\propto \max_{\Gamma \succeq 0} \hat{p}(S;\Gamma) \quad (13) \\
p(X) &\propto \max_{\Psi \succeq 0} \hat{p}(X;\Psi) \quad (14)
\end{aligned}$$

where

$$\begin{aligned}
\hat{p}(S;\Gamma) &\triangleq \exp\left[-\frac{1}{2}\sum_{ij}\left(\frac{s_{ij}^2}{\gamma_{ij}} + \log\gamma_{ij}\right)\right] \quad (15) \\
\hat{p}(X;\Psi) &\triangleq \exp\left[-\frac{1}{2}\text{Trace}\left[XX^T\Psi^{-1}\right] - \frac{n}{2}\log|\Psi|\right],
\end{aligned}$$

which can be viewed as unnormalized approximate priors offering strict lower bounds on $p(S)$ and $p(X)$. We also then obtain a tractable posterior approximation given by

$$\hat{p}(X,S|Y;\Gamma,\Psi) \triangleq \frac{p(Y|S,X)\hat{p}(S;\Gamma)\hat{p}(X;\Psi)}{\int p(Y|S,X)\hat{p}(S;\Gamma)\hat{p}(X;\Psi)dSdX}. \quad (16)$$

Here $\hat{p}(X,S|Y;\Gamma,\Psi)$ is a Gaussian distribution with closed-form first and second moments, e.g., the means of $S$ and $X$ are actually given by the righthand sides of (9). The question remains how to choose $\Gamma$ and $\Psi$. With the goal of reflecting the mass of the true distribution $p(Y,X,S)$, we adopt the approach from Wipf et al. (2011) and attempt to solve

$$\min_{\Psi,\Gamma}\int |p(Y,X,S) - p(Y|S,X)\hat{p}(S;\Gamma)\hat{p}(X;\Psi)|\,dXdS \quad (17)$$

$$= \min_{\Psi,\Gamma}\int p(Y|S,X)\,|p(X)p(S) - \hat{p}(S;\Gamma)\hat{p}(X;\Psi)|\,dXdS. \quad (18)$$

The basic idea here is that we only care that the approximate priors match the true ones in regions where the likelihood function $p(Y|X,S)$ is significant; in other regions the mismatch is more or less irrelevant. Moreover, by virtue of the strict lower variational bound, (18) reduces to

$$\max_{\Psi,\Gamma}\int p(Y|S,X)\hat{p}(S;\Gamma)\hat{p}(X;\Psi)dXdS \equiv \min_{\Psi,\Gamma}\mathcal{L}(\Psi,\Gamma) \quad (19)$$

where

$$\mathcal{L}(\Psi,\Gamma) \triangleq \sum_{j=1}^{n}\left[\boldsymbol{y}_j^T\Sigma_{y_j}^{-1}\boldsymbol{y}_j + \log|\Sigma_{y_j}|\right] \quad (20)$$

with

$$\Sigma_{y_j} \triangleq \Psi + \bar{\Gamma}_j + \lambda I. \quad (21)$$

This $\Sigma_{y_j}$ can be viewed as the covariance of the $j$-th column of $Y$ given fixed values of $\Psi$ and $\Gamma$. To recap then, we need now minimize $\mathcal{L}(\Psi,\Gamma)$ with respect to $\Psi$ and $\Gamma$, and then plug these estimates into (16) giving the approximate posterior. The mean of this distribution (see below) can then be used as a point estimate for $X$ and $S$. This process is sometimes referred to as *empirical Bayes* because we are using the data to guide our search for an optimal prior distribution (Berger, 1985; Tipping, 2001).

### 3.1 UPDATE RULE DERIVATIONS

It turns out that minimization of (20) can be accomplished concurrently with computation of the posterior mean leading to simple, efficient update rules. While (20) is non-convex, we can use a majorization-minimization approach analogous to that used for MAP estimation. For this purpose, we utilize simplifying upper bounds on both terms of the cost function as has been done for related sparse estimation problems Wipf & Nagarajan (2010).

First, the data-dependent term is concave with respect to $\Psi^{-1}$ and $\Gamma^{-1}$ and hence can be expressed as a minimization over $(\Psi^{-1}, \Gamma^{-1})$-dependent hyperplanes. With some linear algebra, it can be shown that

$$\boldsymbol{y}_j^T \Sigma_{y_j}^{-1} \boldsymbol{y}_j = \min_{\boldsymbol{x}_j, \boldsymbol{s}_j} \frac{1}{\lambda} \|\boldsymbol{y}_j - \boldsymbol{x}_j - \boldsymbol{s}_j\|_{\mathcal{F}}^2 + \boldsymbol{x}_j^T \Psi^{-1} \boldsymbol{x}_j + \sum_i \frac{s_{ij}^2}{\gamma_{ij}} \tag{22}$$

for all $j$. With a slight abuse of notation, we adopt $X = [\boldsymbol{x}_1, \ldots, \boldsymbol{x}_n]$ and $S = [\boldsymbol{s}_1, \ldots, \boldsymbol{s}_n]$ as the variational parameters in (22) because they end up playing the same role as the unknown low-rank and sparse coefficients and provide a direct link to the MAP estimates. Additionally, the $\boldsymbol{x}_j$ and $\boldsymbol{s}_j$ which minimize (22) turn out to be equivalent to the posterior means of (16) given $\Psi$ and $\Gamma$ and will serve as our point estimates.

Secondly, for the log-det term, we first use the determinant identity

$$\log |\Psi + \bar{\Gamma}_j + \lambda I| = \log |\Psi| + \log |\bar{\Gamma}_j| + \log |A_j| + C, \tag{23}$$

where

$$A_j \triangleq \lambda^{-1} \begin{bmatrix} I & I \\ I & I \end{bmatrix} + \begin{bmatrix} \Psi^{-1} & \boldsymbol{0} \\ \boldsymbol{0} & \bar{\Gamma}_j^{-1} \end{bmatrix} \tag{24}$$

and $C$ is an irrelevant constant. The term $\log |A_j|$ is jointly concave in both $\Psi^{-1}$ and $\bar{\Gamma}_j^{-1}$ and thus can be bounded in a similar fashion as (22), although a closed-form solution is no longer available. (Other decompositions lead to different bounds and different candidate update rules.) Here we use

$$\log |A_j| = \tag{25}$$
$$\min_{U_j, V_j \succeq 0} \text{Trace} \left[ U_j^T \Psi^{-1} + V_j^T \bar{\Gamma}_j^{-1} \right] - h^*(U_j, V_j)$$

where $h^*(U_j, V_j)$ is the concave conjugate function of $\log |A_j|$ with respect to $\Psi^{-1}$ and $\bar{\Gamma}_i^{-1}$. Note that while $h^*(U_j, V_j)$ has no closed-form solution, the minimizing values of $U_j$ and $V_j$ can be computed in closed-form via

$$U_j = \frac{\partial \log |A_j|}{\partial \Psi^{-1}}, \qquad V_j = \frac{\partial \log |A_j|}{\partial \bar{\Gamma}_j^{-1}}. \tag{26}$$

When we drop the minimizations over the variational parameters $\boldsymbol{x}_j$, $\boldsymbol{s}_j$, $U_j$, and $V_j$ for all $j$, we arrive at a convenient family of upper bounds on the cost function $\mathcal{L}(\Psi, \Gamma)$. Given some estimate of $\Psi$ and $\Gamma$, we can evaluate all variational parameters in closed form (see below). Likewise, given all of the variational parameters we can solve directly for $\Psi$ and $\Gamma$ because now $\mathcal{L}(\Psi, \Gamma)$ has been conveniently decoupled and we need only compute

$$\min_{\Psi \succeq 0} \sum_j \left( \boldsymbol{x}_j^T \Psi^{-1} \boldsymbol{x}_j + \text{Trace} \left[ U_j^T \Psi^{-1} \right] \right) + n \log |\Psi| \tag{27}$$

and

$$\min_{\gamma_{ij} \geq 0} \frac{s_{ij}^2}{\gamma_{ij}} + \frac{[V_j]_{ii}}{\gamma_{ij}} + \log \gamma_{ij}, \quad \forall i, j. \tag{28}$$

We summarize the overall procedure next.

### 3.2 ALGORITHM SUMMARY

1. Compute $\kappa \triangleq \frac{1}{nm} \|Y\|_{\mathcal{F}}^2$

2. Initialize $\Psi^{(0)} \to \kappa I$, and $\bar{\Gamma}_j^{(0)} \to \kappa I$ for all $j$.

3. For the $(k+1)$-th iteration, compute the optimal $\boldsymbol{x}_j$ and $\boldsymbol{s}_j$ via

$$\boldsymbol{x}_j^{(k+1)} \to \Psi^{(k)} \left( \Psi^{(k)} + \bar{\Gamma}_j^{(k)} + \lambda I \right)^{-1} \boldsymbol{y}_j$$
$$\boldsymbol{s}_j^{(k+1)} \to \bar{\Gamma}^{(k)} \left( \Psi^{(k)} + \bar{\Gamma}_j^{(k)} + \lambda I \right)^{-1} \boldsymbol{y}_j \tag{29}$$

4. Likewise, compute the optimal $U_j$ and $V_j$ via

$$U_j^{(k+1)} \to \Psi^{(k)} - \Psi^{(k)} \left( \Psi^{(k)} + \bar{\Gamma}_j^{(k)} + \lambda I \right)^{-1} \Psi^{(k)}$$
$$V_j^{(k+1)} \to \bar{\Gamma}_j^{(k)} - \bar{\Gamma}_j^{(k)} \left( \Psi^{(k)} + \bar{\Gamma}_j^{(k)} + \lambda I \right)^{-1} \bar{\Gamma}_j^{(k)} \tag{30}$$

5. Update $\Psi$ and $\Gamma$ using the new variational parameters via

$$\Psi^{(k+1)} \to \frac{1}{n} \sum_j \left[ \boldsymbol{x}_j^{(k+1)} \left( \boldsymbol{x}_j^{(k+1)} \right)^T + U_j^{(k+1)} \right]$$
$$\gamma_{ij}^{(k+1)} \to \left( s_{ij}^{(k+1)} \right)^2 + \left[ V_j^{(k+1)} \right]_{ii}, \forall i, j \tag{31}$$

6. Repeat steps 3 through 5 until convergence. (Recall that $\bar{\Gamma}_j$ is a diagonal matrix formed from the $j$-th column of $\Gamma$.) This process is guaranteed to reduce or leave unchanged the cost function at each iteration.

Note that if we set $U_j^{(k+1)}, V_j^{(k+1)} \to 0$ for all $j$, then the algorithm above is guaranteed to (at least locally) minimize the MAP cost function from (5). Additionally, for matrix completion problems (Candès & Recht, 2008), where the support of the sparse errors is known a priori, we need only set each $\gamma_{ij}$ corresponding to a corrupted entry to $\infty$. This limiting case can easily be handled with efficient reduced rank updates.

One positive aspect of this algorithm is that it is largely parameter free. We must of course choose some stopping criteria, such as a maximum number of iterations or a convergence tolerance. (For all experiments in Section 5 we simply set the maximum number of iterations at 100.) We must also choose some value for $\lambda$, which balances allowable contributions from a diffuse error matrix $E$, although frequently methods have some version of this parameter, including the PCP algorithm. For all of our experiments we simply choose $\lambda = 10^{-6}$ since we did not include an $E$ component consistent with the original RPCA formulation from Candès et al. (2011).

From a complexity standpoint, each iteration of the above algorithm can be computed in $O(m^3 n)$, where $n \geq m$, so it is linear in the larger dimension of $Y$ and cubic in the smaller dimension. For many computer vision applications (see Section 5 for one example), images are vectorized and then stacked, so $Y$ may be $m =$number-of-images by $n =$ number-of-pixels. This is relatively efficient, since the number of images may be on the order of 100 or fewer (see Wu et al. (2010)). However, when $Y$ is a large square matrix, the updates are more expensive to compute. In the future we plan to investigate various approximation techniques to handle this scenario.

As a final implementation-related point, when given access to a priori knowledge regarding the rank of $X$ and/or sparsity of $S$, it is possible to bias the algorithm's initialization (from Step 1 above) and improve the estimation accuracy. However, we emphasize that for all of the experiments reported in Section 5 we assumed no such knowledge.

### 3.3 Alternative Bayesian Methods

Two other Bayesian-inspired methods have recently been proposed for solving the RPCA problem. The first from Ding et al. (2011) is a hierarchical model with conjugate prior densities on model parameters at each level such that inference can be performed using a Gibbs sampler. This method is useful in that the $\lambda$ parameter balancing the contribution from diffuse errors $E$ is estimated directly from the data. Moreover, the authors report significant improvement over PCP on example problems. A potential downside of this model is that theoretical analysis is difficult because of the underlying complexity. Additionally, a large number of MCMC steps are required to obtain good estimates leading to a significant computational cost even when $Y$ is small. It also uses an estimate of Rank$[X]$ which can effect the convergence rate of the Gibbs sampler.

A second method from Babacan et al. (2011) similarly employs a hierarchial Bayesian model but uses a factorized mean-field variational approximation for inference (Attias, 2000). Note that this is an entirely different type of variational method than ours, relying on a posterior distribution that factorizes over $X$ and $S$, meaning $p(X, S|Y) \approx q(X|Y)q(S|Y)$, where $q(X|Y)$ and $q(S|Y)$ are approximating distributions learned by minimizing a free energy-based cost function.[4] Unlike our model, this factorization implicitly decouples $X$ and $S$ in a manner akin to MAP estimation, and may potentially produce more locally minimizing solutions (see analysis below). Moreover, while this approach also has a mechanism for estimating $\lambda$, there is no comprehensive evidence given that it can robustly expand upon the range of corruptions and rank that can already be handled by PCP.

To summarize both of these methods then, we would argue that while they offer a compelling avenue for computing $\lambda$ automatically, the underlying cost functions are substantially more complex than PCP or our method rendering more formal analyses somewhat difficult. As we shall see in Sections 4 and 5, the empirical Bayesian cost function we propose is analytically principled and advantageous, and empirically outperforms PCP by a wide margin.

## 4 ANALYSIS

In this section we will examine global and local minima properties of the proposed method and highlight potential advantages over MAP, of which PCP can also be interpreted as a special case. For analysis purposes and comparisons with MAP estimation, it is helpful to convert the empirical Bayes cost function (20) into $(X, S)$-space by first optimizing over $U_j$, $V_j$, $\Psi$ and $\Gamma$, leaving only the unknown coefficient matrices $X$ and $S$. Using this process, it is easily shown that the estimates of $X$ and $S$ obtained by globally (or locally) minimizing (20) will also globally (or locally) minimize

$$\min_{X,S} \|Y - X - S\|_\mathcal{F}^2 + \lambda g_{EB}(X, S; \lambda), \quad (32)$$

where the penalty function is given by

$$g_{EB}(X, S; \lambda) \triangleq \quad (33)$$

$$\min_{\Gamma \succeq 0, \Psi \succeq 0} \sum_{i=1}^{n} \boldsymbol{x}_j \Psi^{-1} \boldsymbol{x}_j + \boldsymbol{s}_j^T \bar{\Gamma}_j^{-1} \boldsymbol{s}_j + \log \left| \Psi + \bar{\Gamma}_j + \lambda I \right|.$$

Note that the implicit MAP penalty from (5) is nearly identical:

$$g_{map}(X, S) \triangleq \quad (34)$$

---
[4]Additional factorizations are also included in the model.

$$\min_{\Gamma \succeq 0, \Psi \succeq 0} \sum_{i=1}^{n} \boldsymbol{x}_j \Psi^{-1} \boldsymbol{x}_j + \boldsymbol{s}_j^T \bar{\Gamma}_j^{-1} \boldsymbol{s}_j + \log|\Psi| + \log|\bar{\Gamma}_j|.$$

The primary distinction is that in the MAP case the variational parameters separate whereas in empirical Bayesian case they do not. (Note that, as discussed below, we can apply a small regularizer analogous to $\lambda$ to the log terms in the MAP case as well.) This implies that $g_{map}(X, S)$ can be expressed as some $g_{map}(X) + g_{map}(S)$ whereas $g_{EB}(X, S; \lambda)$ cannot. A related form of non-separability has been shown to be advantageous in the context of sparse estimation from overcomplete dictionaries (Wipf et al., 2011).

We now examine how this crucial distinction can be beneficial in producing maximally sparse, low-rank solutions that optimize (2). We first demonstrate how (32) mimics the global minima profile of (2). Later we show how the smoothing mechanism of the empirical Bayesian marginalization can mitigate spurious locally minimizing solutions.

The original RPCA development from Candès et al. (2011) assumes that $E = 0$, which is somewhat easier to analyze. We consider this scenario first.

**Theorem 1.** Assume that there exists at least one solution to $Y = X+S$ such that $\text{Rank}[X] + \max_j \|\boldsymbol{s}_j\|_0 < m$. Then in the limit as $\lambda \to 0$, any solution that globally minimizes (32) will globally minimize (2).

Proofs will be deferred to a subsequent journal publication. Note that the requirement $\text{Rank}[X] + \max_j \|\boldsymbol{s}_j\|_0 < m$ is a relatively benign assumption, because without it the matrices $X$ and $S$ are formally unidentifiable even if we are able to globally solve (2). For $E > 0$, we may still draw direct comparisons between (32) and (2) when we deviate slightly from the Bayesian development and treat $g_{EB}(X, S; \lambda)$ as an abstract, stand-alone penalty function. In this context we may consider $g_{EB}(X, S; \alpha)$, with $\alpha \neq \lambda$ as a more general candidate for estimating RPCA solutions.

**Corollary 1.** Assume that $X_{(\lambda)}$ and $S_{(\lambda)}$ are a unique, optimal solution to (2) and that $\text{Rank}\left[X_{(\lambda)}\right] + \max_j \|\left[\boldsymbol{s}_{(\lambda)}\right]_j\|_0 < m$. Then there will always exist some $\lambda'$ and $\alpha'$ such that the global minimum of

$$\min_{X,S} \|Y - X - S\|_{\mathcal{F}}^2 + \lambda' g_{EB}(X, S; \alpha'), \quad (35)$$

denoted $X_{(\lambda', \alpha')}$ and $S_{(\lambda', \alpha')}$, satisfies the conditions $\|X_{(\lambda', \alpha')} - X_{(\lambda)}\| < \epsilon$ and $\|S_{(\lambda', \alpha')} - S_{(\lambda)}\| < \epsilon$, where $\epsilon$ can be arbitrarily small.

Of course MAP estimation can satisfy a similar property as Theorem 1 and Corollary 1 after a minor modification. Specifically, we may define

$$g_{map}(X, S; \alpha) \triangleq \qquad (36)$$

$$\min_{\Gamma \succeq 0, \Psi \succeq 0} \sum_{j=1}^{n} \boldsymbol{x}_j \Psi^{-1} \boldsymbol{x}_j + \boldsymbol{s}_j^T \bar{\Gamma}_j^{-1} \boldsymbol{s}_j + \log|\Psi + \alpha| + \log|\bar{\Gamma}_j + \alpha|$$

and then achieve a comparable result to the above using $g_{map}(X, S; \alpha')$. The advantage of empirical Bayes then is not with respect to global minima, but rather with respect to local minima. The separable, additive low-rank plus sparsity penalties that emerge from MAP estimation will always suffer from the following limitation:

**Theorem 2.** Let $S_{ij}^{(a)}$ denote any matrix $S$ with $s_{ij} = a$. Now consider any optimization problem of the form

$$\min_{X,S} g_1(X) + g_2(S), \quad \text{s.t. } Y = X + S, \quad (37)$$

where $g_1$ is an arbitrary function of the singular values of $X$ and $g_2$ is an arbitrary function of the magnitudes of the elements in $S$. Then to ensure that a global minimum of (37) is a global minimum of (2) for all possible $Y$, we require that

$$\lim_{\epsilon \to 0} \frac{g_2\left[S_{ij}^{(\epsilon)}\right] - g_2\left[S_{ij}^{(0)}\right]}{\epsilon} = \infty \quad (38)$$

for all $i$ and $j$ and $S$. An analogous condition holds for the function $g_1$.

This result implies that whenever an element of $S$ approaches zero, it will require increasing the associated penalty $g_2(S)$ against an arbitrarily large gradient to escape in cases where this coefficient was incorrectly pruned. Likewise, if the rank of $X$ is prematurely reduced in the wrong subspace, there may be no chance to ever recover since this could require increasing $g_1(X)$ against an arbitrarily large gradient factor. In general, Theorem 2 stipulates that if we would like to retain the same global minimum as (2) using a MAP estimation-based cost function, then we will necessarily enter an inescapable basin of attraction whenever *either* $\text{Rank}[X] < m$ or $\|\boldsymbol{s}_j\|_0 < m$ for some $j$. This is indeed a heavy price to pay.

Crucially, because of the coupling of low-rank and sparsity regularizers, the penalty function $g_{EB}(X, S; \lambda)$ does not have this limitation. In fact, we only encounter insurmountable gradient barriers when $\text{Rank}[X] + \|\boldsymbol{s}_j\|_0 < m$ for some $j$, in which case the covariance $\Sigma_{y_j}$ from (21) becomes degenerate (with $\lambda$ small), a much weaker condition. To summarize (emphasize) this point then, MAP can be viewed as heavily dependent on degeneracy of the matrices $\Psi$ and $\Gamma$ in isolation, whereas empirical Bayes is only sensitive to degeneracy of their summation.

This distinction can also be observed in how the effective penalties on $X$ and $S$, as imbedded in $g_{EB}(X, S; \lambda)$, vary given fixed values of $\Gamma$ or $\Psi$ respectively. For example, when $\Psi$ is close to being full rank and orthogonal (such as when the algorithm is initialized), then the implicit penalty on $S$ is minimally non-convex (only slightly concave). In fact, as $\Psi$ becomes large and orthogonal, the penalty converges to a scaled version of the $\ell_1$ norm. In contrast, as $\Psi$ becomes smaller and low-rank, the penalty approaches a scaled version of the $\ell_0$ norm, implying that maximally sparse corruptions will be favored. Thus, we do not aggressively favor maximally sparse $S$ until the rank has already been reduced and we are in the basin of attraction of a good solution. Of course no heuristic annealing strategy is necessary, the transition is handled automatically by the algorithm.

Additionally, whenever $\Psi$ is fixed, the resulting cost function formally decouples into $n$ separate, canonical sparse estimation problems on each $s_j$ in isolation. With $\lambda = 0$, it not difficult to show that each of these subproblems is equivalent to solving

$$\min_{s_j} \|y_j - \Phi s_j\|_2^2 + g_{EB}(s_j) \quad (39)$$

where

$$g_{EB}(s_j) \triangleq \min_{\gamma_j \geq 0} \sum_{j=1}^{n} s_j^T \bar{\Gamma}_j^{-1} s_j + \log \left| \Phi \bar{\Gamma}_j \Phi^T + I \right| \quad (40)$$

is a concave sparsity penalty on $s_j$ and $\Phi$ is any matrix such that $\Phi \Psi \Phi^T = I$.[5] When $\Phi$ is nearly orthogonal, this problem has no local minima and a global solution that approximates the hard thresholding of the $\ell_0$ norm; however, direct minimization of the $\ell_0$ norm will have $2^n$ local minima (Wipf et al., 2011). In contrast, when $\Phi$ is poorly conditioned (with approximately low-rank structure, it has been argued in Wipf et al. (2011) that penalties such as $g_{EB}(s_j)$ are particularly appropriate for avoiding local minima.

Something similar occurs when $\Gamma$ is now fixed and we evaluate the penalty on $X$. This penalty approaches something like a scaled version of the nuclear norm (less concave) when elements of $\Gamma$ are set to a large constant and it behaves more like the rank function when $\Gamma$ is small. At initialization, when $\Gamma$ is all ones, we are relatively free to move between solutions of various rank without incurring a heavy penalty. Later as $\Gamma$ becomes sparse, solutions satisfying $\text{Rank}[X] + \|s_j\|_0 < m$ for some $j$ become heavily favored.

---

[5]We have assumed here that $\Psi$ is full rank.

As a final point, the proposed empirical Bayesian approach can be implemented with alternative variational bounds and possibly optimized with something akin to simultaneous reweighted nuclear and $\ell_1$ norm minimization, a perspective that naturally suggests further performance analyses such as those applied to sparse estimation in Wipf & Nagarajan (2010).

## 5 EMPIRICAL RESULTS

This section provides some empirical evidence for the efficacy of our RPCA method. First, we present comparisons with PCP recovering random subspaces from corrupted measurements. Later we discuss a photometric stereo application. In all cases we used the the augmented lagrangian method (ALM) from Lin et al. (2010) to implement PCP. This algorithm has efficient, guaranteed convergence and in previous empirical tests ALM has outperformed a variety of other methods in computing minimum nuclear norm plus $\ell_1$ norm solutions.

### 5.1 RANDOM SUBSPACE SIMULATIONS

Here we demonstrate that the empirical Bayesian algorithm from Section 3.2, which we will refer to as EB, can recovery unknown subspaces from corrupted measurements in a much broader range of operating conditions compared to the convex PCP. In particular, for a given value of $\text{Rank}[X]$, our method can handle a substantially larger fraction of corruptions as measured by $\rho = \|S\|_0/(nm)$. Likewise, for a given value of $\rho$, we can accurately estimate an $X$ with much higher rank. Consistent with Candès et al. (2011), we consider the case where $E = 0$, such that all the error is modeled by $S$. This allows us to use the stable, convergent ALM code available online.[6]

The first experiment proceeds as follows. We generate a low-rank matrix $X$ with dimensions reflective of many computer vision problems: *number-of-images × number-of-pixels*. Here we choose $m = 20$ and $n = 10^4$, the later dimension equivalent to a $100 \times 100$ pixel image. For each trial, we compute an $m \times n$ matrix with iid $\mathcal{N}(0, 1)$ entries. We then compute the SVD of this matrix and set all but the $r$ largest singular values to zero to produce a low-rank $X$. $S$ is generated with nonzero entries selected uniformly with probability $\rho = 0.2$. Nonzero values are sampled from an iid Uniform[-10,10] distribution. We then compute $Y = X + S$ and try to estimate $X$ and $S$ using the EB and PCP algorithms. Estimation results averaged over multiple trials as $r$ is varied from 1 to 10 are depicted in Figure 1. We plot normalized mean-squared error

---

[6]http://perception.csl.uiuc.edu/matrix-rank/

(MSE) as computed via $\left\langle \|X - \hat{X}\|_{\mathcal{F}}^2 / \|X\|_{\mathcal{F}}^2 \right\rangle$ as well as the average angular error between the estimated and true subspaces. In both cases the average is across 10 trials.

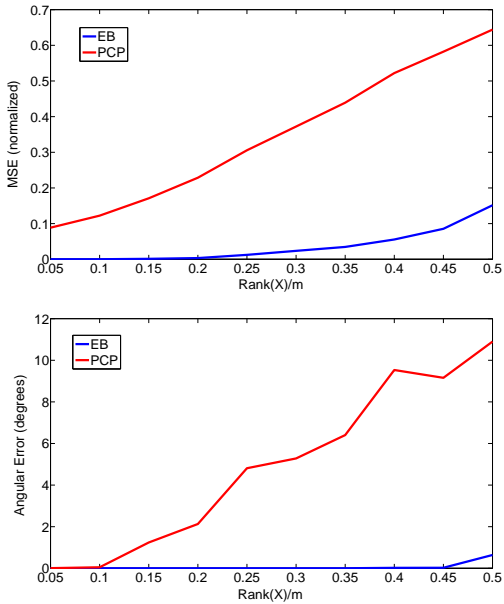

Figure 1: Estimation results where the corruption probability $\rho$ is 0.2 and $\text{Rank}[X]/m$ is varied from 0.05 to 0.50, meaning up to 50% of full rank. *Top*: Normalized mean-squared error (MSE). *Bottom*: Average angle (in degrees) between the estimated and true subspaces.

From Figure 1 we observe that EB can accurately estimate $X$ for substantially higher values of the rank. Interestingly, we are also still able to estimate the correct subspace spanned by columns of $X$ perfectly even when the MSE of estimating $X$ starts to rise (compare Figure 1(*Top*) with Figure 1(*Bottom*)). Basically, this occurs because, even if we have estimated the subspace perfectly, reducing the MSE to zero implicitly requires solving a challenging sparse estimation problem for every observation column $y_j$. For each column, this problem requires learning $d_j \triangleq \text{Rank}[X] + \|s_j\|_0$ nonzero entries given only $m = 20$ observations. For our experiment, we can have $d_j > 14$ with high probability for some columns when the rank is high, and thus we may expect some errors in $\hat{S}$ (not shown). However, the encouraging evidence here is that EB is able to keep these corrupting errors at a minimum and estimate the subspace accurately long after PCP has failed. Moreover, if an accurate estimate of $X$ is needed, as opposed to just the correct spanning subspace, then a postprocessing error correction step can potentially be applied to each column individually to improve performance.

The second experiment is similar to the first only now we hold $\text{Rank}[X]$ fixed at 4, meaning $\text{Rank}[X]/m = 0.2$, and vary the fraction of corrupted entries in $S$ from 0.1 to 0.8. Figure 2 shows that EB is again able to drastically expand the range whereby successful estimates are obtained. Notably it is able to recover the correct subspace even with 70% corrupted entries.

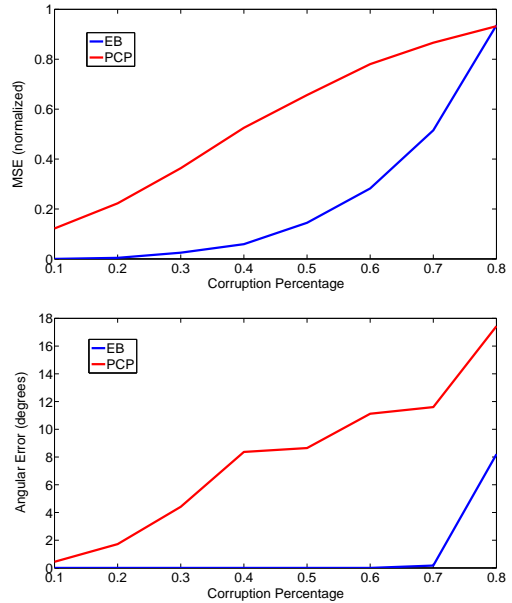

Figure 2: Estimation results with $\text{Rank}[X]/m = 0.2$ and $\rho$ varied from 0.1 to 0.8. *Top*: Normalized mean-squared error (MSE). *Bottom*: Average angle (in degrees) between the estimated and true subspaces.

As a final comparison, we tested PCP and EB on a $400 \times 400$ observation matrix $Y$ generated as above with a $\text{Rank}[X]/m = 0.1$ and $\rho = 0.5$. The estimation results are reported in Table 1. PCP performs poorly since the normalized MSE is above one, meaning we would have been better off simply choosing $\hat{X} = 0$ in this regard. Additionally, the angular error is very near 90 degrees, consistent with the error from a randomly chosen subspace in high dimensions. In contrast, EB provides a reasonably good estimate considering the difficulty of the problem.

Table 1: Estimation results with $m = n = 400$, $\text{Rank}[X]/m = 0.1$ and $\rho = 0.5$.

|  | **PCP** | **EB** |
| --- | --- | --- |
| MSE (norm.) | 1.235 | 0.066 |
| Angular Error | 88.50 | 5.01 |

## 5.2 PHOTOMETRIC STEREO

Photometric stereo is a method for estimating surface normals of an object or scene by capturing multiple images from a fixed viewpoint under different lighting conditions (Woodham, 1980). At a basic level, this methodology assumes a Lambertian object surface, point light sources at infinity, an orthographic camera view, and a linear sensor response function. Under these conditions, it has been shown that the intensities of a vectorized stack of images $Y$ can be expressed as

$$Y = L^T N \Upsilon, \qquad (41)$$

where $L$ is a $3 \times m$ matrix of $m$ normalized lighting directions, $N$ is a $3 \times n$ matrix of surface normals at $n$ pixel locations, and $\Upsilon$ is a diagonal matrix of diffuse albedo values (Woodham, 1980). Thus, if we were to capture at least 3 images with known, linearly independent lighting directions we can solve for $N$ using least squares. Of course in reality many common non-Lambertian effects can disrupt this process, such as specularities, cast or attached shadows, and image noise, etc. In many cases, these effects can be modeled as an additive sparse error term $S$ applied to (41).

As proposed in Wu et al. (2010), we can estimate the subspace containing $N$ by solving (2) assuming $X = L^T N \Upsilon$ and $E = 0$. The resulting $\hat{X}$, combined with possibly other *a priori* information regarding the lighting directions $L$ can lead to an estimate of $N$. Wu et al. (2010) propose using a modified version of PCP for this task, where a shadow mask is included to simplify the sparse error correction problem. However, in practical situations it may not always be possible to accurately locate all shadow regions in this manner so it is desirable to treat them as unknown sparse corruptions.

For this experiment we consider the synthetic Caesar image from the INRIA 3D Meshes Research Database[7] with known surface normals. Multiple 2D images with different known lighting conditions can easily be generated using the Cook-Torrance reflectance model (Cook & Torrance, 1981). These images are then stacked to produce $Y$. Because shadows are extremely difficult to handle in general, as a preprocessing step we remove rows of $Y$ corresponding to pixel locations with more than 10% shadow coverage. Specular corruptions were left unfiltered. We tested our algorithm as the number of images, drawn randomly from a batch of 40 total, was varied from 10 to 40. Results averaged across 5 trials are presented in Figure 3. The error metrics have been redefined to accommodate the photometric stereo problem. We now define the normalized MSE as

---

[7] http://www-roc.inria.fr/gamma/gamma/download/download.php

$\left\langle \|X - \hat{X}\|_{\mathcal{F}}^2 / \|X - Y\|_{\mathcal{F}}^2 \right\rangle$, which measures how much improvement we obtain beyond just using the observation matrix $Y$ directly. Similarly we normalized the angular error by dividing by the angle between $Y$ and the true $X$ for each trial.

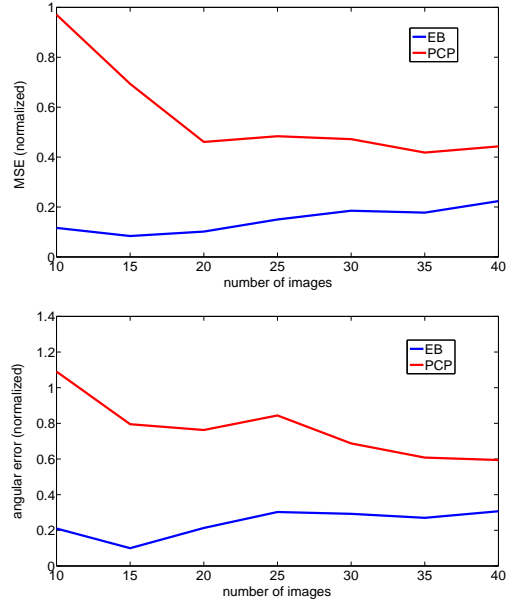

Figure 3: Estimation results using photometric stereo data as the number of images $m$ is varied from 10 to 40. *Top*: Normalized MSE (see altered definition in the main text). *Bottom*: Normalized angle between the estimated and the true subspaces.

From Figure 3 it is clear that EB outperforms PCP in both MSE and angular error, especially when there are fewer images present. It is not entirely clear however why the MSE and angular error are relatively flat for EB as opposed to dropping lower as $m$ increases. Of course these are errors relative to using $Y$ directly to predict $X$, which could play a role in this counterintuitive effect.

## 6 CONCLUSIONS

In this paper we have analyzed a new empirical Bayesian approach for matrix rank minimization in the context of RPCA, where the goal is to decompose a given data matrix into low-rank and sparse components. Using a variational approximation and subsequent marginalization, we ultimately arrive at a novel regularization term that couples low-rank and sparsity penalties in such a way that locally minimizing solutions are effectively smoothed while the global optimum matches that of the ideal RPCA cost function.